# Which symbol grounding problem should we try to solve?

*Vincent C. Müller*

Anatolia College/ACT & University of Oxford
www.sophia.de

21st October, 2013

Floridi and Taddeo propose a condition of "zero semantic commitment" for solutions to the grounding problem, and a solution to it. I argue briefly that their condition cannot be fulfilled, not even by their own solution. After a look at Luc Steel's very different competing suggestion, I suggest that we need to re-think what the problem is and what role the 'goals' in a system play in formulating the problem. On the basis of a proper (syntactic) understanding of computing, I come to the conclusion that the only sensible grounding problem is how we can explain and re-produce the *behavioral ability* and *function* of meaning in artificial computational agents.

## 1. A traditional symbol grounding problem

The traditional symbol grounding problem suggests that mere syntactic symbol manipulation in a computer will never lead to meaning – notoriously formulated in John Searle's 'Chinese Room argument' (Searle, 1980). So, how can the symbols in a computational system be 'grounded' such that they acquire that causal link to their referents?

Searle in the Chinese room says: I do computation. This is a matter of function (computation), not of substance (silicon vs. carbon). He then goes on to argue that:

1) The symbol manipulator in the Chinese room does not *understand* Chinese, but just produces correct output; he has no chance of learning Chinese

2) The whole system: central manipulator + manipulation manuals + input-output system does not understand Chinese, not even if "sensory organs" (cameras, microphones, etc.) are added. All of this supplies "just more Chinese."

*But:* 2) does not follow from 1), as many people have pointed out. This does not mean that his argument fails, however. The upshot of the argument is, in my view, that Searle sets the task to explain how a system can understand Chinese given that the central symbol manipulator does not (Müller, 2009). In other words, the main argument is really this:

1. Syntactical manipulation is not sufficient for a system to acquire meaning.
2. A computer does only syntactical manipulation.
3. A computer will not acquire meaning.

Searle tries to show that a system could not "think, understand and so on *solely* in virtue of being a computer with the right sort of program" (Searle, 1980, pp. §6, 368).

The traditional challenge that Stevan Harnad has made of this is: How can a system (natural or artificial) acquire meaning for their symbols given that the central symbol manipulator is merely syntactical?

The resulting situation is a standoff. The philosophers typically say to the AI researchers:

"Our a priori problem is unsolved, so you will not succeed."

To which the AI researchers respond:

"We couldn't care less and we are doing just fine. (We have a refutation, and we might write it up when we find the time.)"

Of course, this is rather unsatisfactory, so a clarification of the problem like the one offered by Floridi is very welcome. We will investigate Floridi's problem, an alternative proposal by Luc Steels (not discussed by Floridi) and offer our own proposal for a formulation of the problem.

## 2. Zero semantic commitment

Grounding is a problem for the philosophy of information since it needs to make the move from mere 'data' to meaningful information – thus the chapters 6 and 7 in (Floridi, 2011), based on the papers (Taddeo & Floridi, 2005, 2007). In the following, I will refer to the later book version.

The problem in Floridi is first explicitly detached from the discussions around the Chinese Room and then a proposal for "zero semantic commitment" is made, which is a conjunction of three conditions, jointly called the "Z condition":

> "a. No form of innatism is allowed; no semantic resources … should be magically presupposed […]
>
> b. No form of externalism is allowed either: no semantic resources should be uploaded from the 'outside' by some deus ex machina already semantically proficient […]
>
> c. The AA [artificial agent] may have its own capacities and resources" (Floridi, 2011, pp. 137-139)

### 2.1. Semantics & Goals

The Z condition immediately strikes me as very strong. One thing that appears a minimal requirement (a necessary condition) for semantics is directedness, goal orientation of the agent. This looks like a problem for the Z condition, which seems to say that an agent cannot be born with directedness (or be externally supplied with it), since directedness is a necessary condition for semantics. But note that directedness is *not* also a sufficient condition; so if we do not have semantics (yet), nothing follows on whether we also have directedness in the agent as well – but Floridi will make it look like we can't even have directedness. What we really need to know now is what the agent has that allows them to find a 'praxical' solution – something that the mere syntactic system of Searle etc. was lacking.

If, for example, we consider an animal that is trying to survive by finding food and shelter, such an animal might develop interaction with other members of its species that involves symbols for matters they care about. But notice how artificial this consideration is, since we are not assuming such a thing could happen within a generation, ontogenetically (unless we give in to 'Chinese Room' fiction of an agent that already has full blown cognitive ability and now tries to develop semantics). So, one real question is how this can happen in a species – but this is precisely what the Z condition seems to exclude, since phylogenetic developments involve innateness and external semantic resources.

### 2.2. Action-based semantics to the rescue

Let us see what actually happens when a solution is brought to the table. Floridi proposes in his 'action-based semantics' "the proto-meanings of the symbols generated by an AA are the internal states of that AA, which in turn are directly correlated to the actions performed by that same AA" (Floridi, 2011, p. 164). In the example of wall-following of a simple device where a sensor is appropriately linked to actuators (as in a Breitenberg vehicle) the "purpose of the action has no direct influence on the generation of the meaning. No teleosemantics of any sort is presupposed. Hence, in AbS there are no extrinsic semantic criteria driving the process of meaning generation." (Floridi, 2011, p. 165). And yet, we are told that there is meaning, eventually, unlike in the heliotropic behavior of sunflowers. When evaluating this, first note the slide from 'proto-meaning' to 'meaning' here and then how the claim that this device satisfies the Z condition is defended: It has no goals that it is trying to achieve, neither built-in nor supplied externally. This was a complaint earlier, with competing systems, e.g. with Kismet ((2011, p. 158) and Rosenstein and Cohen's 'Ros' ((2011, p. 154) – both systems have something built-in that makes certain data stand out with respect to others, worth to maintain, marking 'success'. Other systems (e.g. Cangelosi & Riga, 2006) use a teacher or competent language user to mark 'success'. The move from *descriptive* to *prescriptive language* is crucial here: when one solution is *better* than another (more successful), then Floridi steps in and says the Z condition has been violated. I agree.

This brings us back to a dilemma: Either the action-based semantics system is successfully building something or it is just like the sunflowers – with "no semantic resources at all" (p. 164), with no meaning or proto-meaning. Without goals, there is no 'trying', nothing is 'better', and there is no 'success'. Either the system really is an agent (Floridi speaks about an "artificial agent", an "AA"), which implies having goals, or it is just a system that interacts with its environment – without goals.

We are told that the two-machine AA with its two level architecture (object-level and meta-level) is "able to associate symbols to the actions that it performs"

(p. 166), but under what conditions would it actually *do* this? What would make that association a symbol?

It appears that having a semantics implies normativity, since without normativity we do not have the 'right or wrong' use of symbols, the 'failure or success' or referring, etc. Note, however, that the reverse does not hold, since even a purely syntactic system seems to require normativity: the three elementary notions of syntactical systems are that of the 'well-formed formula', 'rules of derivation' and 'axioms'. Also, there are natural systems that have goals and thus normativity but no semantics – most biological systems are like that, e.g. plants, protozoa, etc.

We seem thrown back to fairly elementary questions like what is that 'meaning' or 'semantics' that the system is supposed to ground and why should it do such a thing. What is the problem we are trying to solve? In order to get a better grip on this problem, allow me a look at a competing proposal.

### 3. Steels cutting the Gordian knot *(nomen est omen)*

Luc Steels' paper, 'The symbol grounding problem has been solved, so what's next?' (Steels, 2008) states at the outset that "almost 25 years of philosophical discussion have shed little light on the issue... However, I believe that sufficient progress has been made in cognitive science and AI so that we can say that the symbol grounding problem has been solved" (Steels, 2008, p. 223). He declines any invitation to participate in the philosophers' game of disentangling the problem, but instead wields a large sword to cut through the knot. *Nomen est omen*, when 'Steels' swings the blade, and a few heads might be rolling too, if in the way.

His sword is a particular version of interactive robotics. Is it sharp enough and does it cut the knot?

Here is what he says: "Let me return now to the question originally posed by Searle (1980): can a robot deal with grounded symbols? More precisely, is it possible to build an artificial system that has a body, sensors and actuators, signal and image processing, pattern recognition processes, and information structures to store and use semiotic networks [btw. objects, symbols and concepts], and uses all that for communicating about the world or representing information about the world." (226) – Of course, at this point, the philosopher responds with a self-assured smile and begins to explain that this was *not* Searle's original question. But is this philological correction sufficient?

"A first prerequisite for solving the symbol grounding problem is that we can work with physically embodied autonomous agents, ... they move and behave without any remote control or further human intervention once the experiment starts." (236) - [I.e. we do not put in the meaning – though we put in abilities.]

This seems right, we now have all the fashionable items: situated, embodied, perception-motor coupled, interactive. What else do we want?

Is this not pretty much what one of my former selves had demanded for natural (human) grounding? "Our claim is that the causal connection between the nonconceptual content of the object-files and the world provides causal chains that solve the grounding problem and overcome some of the problems associated with causal accounts of reference." (Raftopoulos & Müller, 2006, p. 254)

### 3.1. Cutting the knot is not disentangling it (the philosophers' temptation)

We should point out that this interactive robot is just a 'purely syntactic device' like the ones Searle is attacking. There is no indication how 'understanding' should arise. The aim is externally defined 'communicative success' – 'success' to whom? (Note the implied normativity.) The success is achieved by the 'right' settings of the devices by the programmers – so it is externally introduced. We should admit that the system partially models the function of linguistic communication, but a model of lactation does not produce milk! So, it does not look like we made any progress at all.

### 4. Summary: Grounding pseudo-problems

It appears that there are several problems here that are 'grounding problems' and not all of them deserve philosophical attention or engineering solutions. Allow me to list a few, not aiming for completeness:

1. If cognition is computation over symbols, then we need grounded symbols before we can start.
2. We want a system that generates meaningful symbols ('semantics') but has no goals, neither innate nor external

3. We want to find the right settings for the right behavior that links linguistic symbols to object types
4. We want to understand how some species could develop semantics (phylogenetically)
5. We want to make a system with meaning in it, on the basis of a syntactic computation device with sensors and actuators

I would suggest that only the last two of these problems can be solved – but not by philosophers. Perhaps there is really a hard and an easy problem here. Let me explain:

**5. The hard and the easy problem of symbol grounding**

**5.1. D. Chalmers's two problems of consciousness**

Chalmers famously proposed that there are two problems of consciousness, namely:

> *Hard Problem:* Why and how does physics give rise to conscious experience (to phenomenal consciousness, to 'what it is like')? (Chalmers, 1995; cf. Chalmers, 1996)

> *Easy Problem:* "explanation of cognitive *abilities* and *functions*" of awareness (the ability to discriminate, integrate information, report mental states, focus attention, etc) – computational or neural mechanisms.

As Chalmers adds: "What makes the hard problem hard and almost unique is that it goes beyond problems about the performance of functions. ... *Why is the performance of these functions accompanied by experience?*" (Chalmers, 1995, p. sect. 3)

I want to suggest that there is an analogous set of problems in grounding.

**5.2. The easy problem of symbol grounding**

"How can we explain and re-produce the *behavioral ability* and *function* of meaning [and other intentional phenomena] in artificial computational agents?" To this problem, we already have some answers:
– Classical AI: Any way that works is a good way (ignoring function).
– Steels: We did it! (The suitably constructed computational mechanism acquires a semantic network in interaction with other such mechanisms.)

- E. DiPaolo (Di Paolo, Rohde, & De Jeogher, 2010): None of these systems have any intentional states, desires. They don't have a life! (A precarious one.) Thus, they do not have the right functional architecture, the right causal connections.

### 5.3. The hard problem of symbol grounding

"How does physics give rise to meaning [and other intentional phenomena]?"

A solution to this problem requires behavioral ability in a functional architecture and the right kind of inner mechanism. The *experience* of understanding is an elementary part of what we call 'understanding'. (For Searle, to mean what I say is trying to get someone else to recognize my intentions.) The hard problem directly involves conscious experience, i.e. it involves solving the hard problem of consciousness and might be an ill-formed problem if the respective problem of consciousness is one. - This problem is untouched by evolutionary robotics.

### 6. Back to the easy problem!

The only problem I see here that has any urgency is the 'easy problem':

"How can we explain and re-produce the *behavioral ability* and *function* of meaning [and other intentional phenomena] in artificial [mainly] computational agents?"

We should get the best cognitive science and go for it!

My suspicion is that Floridi is not tackling the easy problem, but I am not clear why he is tackling the very hard problem characterized by the Z condition.